\documentclass{article}
\usepackage{acra}

\usepackage{graphicx}
\usepackage{booktabs}
\usepackage{multirow}
\usepackage{amsmath} % assumes amsmath package installed
\usepackage{amssymb} % assumes amsmath package installed
\usepackage{algpseudocode}
\usepackage{algorithm}
\usepackage{textcomp, gensymb}
\usepackage{cleveref}

\title{Rendering Stable Features Improves Sampling-Based Localisation with Neural Radiance Fields}
\author{Boxuan Zhang, Lindsay Kleeman, Michael Burke\\ Monash University, Australia \\ \{boxuan.zhang, lindsay.kleeman, michael.g.burke\}@monash.edu}

\begin{document}

\maketitle

\begin{abstract}
Neural radiance fields (NeRFs) are a powerful tool for implicit scene representations, allowing for differentiable rendering and the ability to make predictions about unseen viewpoints. There has been growing interest in object and scene-based localisation using NeRFs, with a number of recent works relying on sampling-based or Monte-Carlo localisation schemes. Unfortunately, these can be extremely computationally expensive, requiring multiple network forward passes to infer camera or object pose. To alleviate this, a variety of sampling strategies have been applied, many relying on keypoint recognition techniques from classical computer vision. This work conducts a systematic empirical comparison of these approaches and shows that in contrast to conventional feature matching approaches for geometry-based localisation, sampling-based localisation using NeRFs benefits significantly from \textit{stable} features. Results show that rendering stable features provides significantly better estimation with a tenfold reduction in the number of forward passes required.
\end{abstract}

\section{Introduction}
\label{sec:intro}
The use of implicit scene representations has surged in computer vision and robotics in recent years, driven in large part by the hugely popular Neural Radiance Field (NeRF) \cite{mildenhall2020nerf}. In short, a NeRF captures the 3D structure and radiance of a scene by mapping points in a defined 3D space to the occupancy and view direction dependent colour at these points. This allows image generation through ray tracing based on a camera pose and viewing direction.

The NeRF has stimulated research in various directions \cite{xie2022nerfsurvey}, from enhancing the network capabilities \cite{wang2021nerf--,pumarola2021dnerf,mueller2022instant,tancik2022blocknerf,zhang2024expsupernerf} to apply it in compositional scene understanding for robot manipulation \cite{driess2022learning,kerr2022evo}. There has also been growing interest in applying this new technique to the field of robot localisation and navigation \cite{zhu2022niceslam,Sucar2021imap,nerf-nav,maggio2022locnerf,lin2023pnerf}.

The problem of how to allow a robot to map and localise itself in a given space has been a long-standing research question in the field of computer vision and robotics. A key challenge this question aims to address is how we can localise a robot or camera quickly, accurately and computationally efficiently. Implicit scene representations such as NeRFs has the significant advantage of storing the environment information much more efficiently compared to many other mapping methods such as point cloud based or mesh based methods. This brings potential of new methods to solve the navigation problem, with a challenge -- when the map is stored implicitly, how should we solve the inverse problem of identifying the camera pose that results in a given image?

An immediate option is a gradient-based approach that back-propagates the error between an observed image and a rendered image with respect to the camera pose as proposed in \cite{yen2020inerf}, but this requires a good initial guess for rapid convergence. Given an implicit scene representation, it is not clear how to find this initial guess. As an alternative, sampling-based approaches \cite{tian2023multi,lin2023pnerf} use multiple pose candidates to reconstruct scenes, iteratively re-rendering scenes and selecting good proposals until convergence. These strategies follow a Monte Carlo or sampling-based localisation \cite{fox1999monte} approach.

The drawback of using these techniques with a NeRF, however, is that rendering an image requires a large amount of computational power. For an image of resolution $H \times W$, $H \times W \times N_{pts}$ network forward passes are needed to render the entire image, with $N_{pts}$ the number of 3D point colours and occupancies predicted for ray tracing. For example, even rendering a small $400 \times 400$ image requires as many as 10 million forward passes. This makes full image rendering infeasible for localisation, and hence many works have proposed strategies to reduce this computational expense \cite{Sucar2021imap,yen2020inerf}. Unfortunately, there has been little systematic evaluation of the choices made for these strategies. A core contribution of this paper is a systematic empirical comparison of approaches to help speed up sampling-based localisation techniques using NeRFs.

We explore a range of sampling strategies, varying points along the ray (a key factor affecting rendering quality), particle numbers, point-wise vs patch-based likelihoods, and pixel sampling approaches on both object-oriented synthetic datasets and forward-facing real scenes. A key finding of this work is that, in contrast to conventional wisdom in computer vision where distinct keypoints, such as corners and edges, are often good feature choices, rendering \textit{stable regions}, where the area in the image has low frequency spatial changes improves sampling-based localisation with NeRFs.

\section{Related Work}
\label{sec:related work}
The conventional approach to solve the robot navigation problem relies on Simultaneous Localisation and Mapping (SLAM) \cite{durrant2006slamI,beiley2006slamII}. With the emergence of the NeRF, several works have proposed pipelines to integrate with SLAM. iMap \cite{Sucar2021imap} appears to be the first to apply NeRF-like MLPs to the SLAM problem, achieving tracking at 10 Hz and global map updating at 2 Hz. The iMap pipeline is able to continuously train an MLP, with key frames detected and added to training datasets during the video feed. This network jointly updates both network weights and the camera pose, building on the approach proposed in NeRF-{}- \cite{wang2021nerf--}. To reduce the computational cost of rendering full images, iMap starts by selecting a large number of random pixels to render and then proceeds  to render pixels from this set that are more informative for 3D reconstruction (larger errors).

NICE-SLAM \cite{zhu2022niceslam} and NICER-SLAM \cite{zhu2023nicerslam}, move away from using the NeRF MLP to infer 3D structure, and instead use several decoder networks to update separate feature grid maps to perform 3D reconstruction. The grid maps are simultaneously decoded and compared with the video feed to perform localisation. In both implementations, the entire image is rendered by the pipeline.

The approaches above consider full SLAM solutions, but there has also been work on the localisation problem directly, assuming a map of the environment is known beforehand. For a neural scene representation setting, this means the NeRF network is already trained before performing localisation. iNeRF \cite{yen2020inerf} proposes to perform gradient descent based camera pose estimation in known scenes, achieving camera tracking at 1Hz. Scene images are rendered using a pre-trained NeRF, based on the estimated camera pose, and the error between the rendered image and the observed image is back propagated to update the camera pose. In order to reduce computational requirements for the gradient descent process, selected pixels are rendered to generate the photometric loss (mean squared error between pixels of interest). iNeRF compares random pixel sampling to interest point sampling (using corner detectors in the observed frame) and interest region sampling (drawing pixels from a region around corner points in the observed frame). Their results indicate that pose optimisation using interest points struggles with local minima, that random sampling required large numbers of pixels to be successful, and that region-based sampling is most effective, allowing pose estimation in around 10 iterations.

Our work is closely related to iNeRF, but instead focuses on sampling-based localisation strategies. We benchmark against the random sampling, interest point detection and region-like rendering approaches used in iNeRF. Our results corroborate the region-based findings of iNeRF, but go further to show the importance of rendering \textit{stable} image features when using mean squared error reconstruction losses for localisation.

Interest point rendering has also been used in other NeRF navigation pipelines. Nerf-navigation \cite{nerf-nav} enables trajectory planning and optimization for robots navigating in a known environment represented by a trained NeRF network. Here, robot state estimation relies on comparing an observed and rendered image at an estimated pose, with an ORB detector selecting areas to render to reduce compute.

The techniques above have mostly considered gradient-based or local optimisation strategies requiring good initial pose estimates. There have also been a number of sampling-based approaches to NeRF localisation. Loc-NeRF \cite{maggio2022locnerf} performs 6DoF robot rover localisation with vision-based perception, following a standard Monte Carlo localisation (MCL) \cite{fox1999monte} implementation and was able to localise a mobile robot in a lab environment with small error. However, this approach relies on a large number of NeRF forward passes (over 70 million), even with randomly selected pixel rendering. Sampling-based localisation strategies (Covariance Matrix Adaptation Evolution Strategy -- CMA-ES) have also been used to locate objects and infer lighting in compositional scene rendering problems \cite{tian2023multi}, but this is extremely slow, with initial pose estimation on the order of 30 minutes. It is clear that more effective strategies are needed for sampling-based localisation using NeRFs. Our work seeks to address this challenge by systematically running experiments over various ablation studies.

It is also important to note that more recent 3D Gaussian Splatting (3DGS)  techniques \cite{kerbl3Dgaussians} approach image rendering in a different way and related applications \cite{Lee_2024_CVPR} have shown much faster training and image rendering than NeRFs. \cite{chen2024safersplat} has shown the potential in localisation and navigation applications. However, regardless of the rendering model used, when it comes to localisation from rendered scenes, there is still significant value in reducing the number of rendering calls required.

\section{Preliminaries}
\label{sec:prelm}

\subsection{Neural Radiance Fields - NeRF}
\label{sec:nerf}

As mentioned before, a NeRF is a simple MLP that can store the 3D information and radiance of a given object or scene in its weights. A NeRF is trained to predict this information within defined boundaries, using images of a given scene along with the corresponding camera poses from which these images were taken. This allows for images to be viewed from arbitrary poses, by querying the network at points along rays intersecting with each pixel of an image at a desired viewpoint.

Following \cite{mildenhall2020nerf}, a ray to be rendered starts from the camera origin $\mathbf{o}$, passes through a pixel of interest on the image plane, and stops at a pre-defined boundary. The direction of the ray is defined by the vector $\mathbf{d}$. Colour estimation is performed by querying the NeRF at points along the ray. We assume that these points are uniformly sampled along the ray, with each point is represented by the vector $\mathbf{r} \left( t \right)=\mathbf{o}+t\mathbf{d}$, with $t \in (t_{near}, t_{far})$. Note that the two vectors $\mathbf{o}$ and $\mathbf{d}$ are calculated based on the pose of the camera in the world coordinate $T_{cam}$ and a known camera intrinsic matrix $K$. The number of points to be sampled along the ray is denoted as $N_{samples}$.

The NeRF predicts the RGB colour $\mathbf{c} \left( \mathbf{r} \left( t \right) \right)$ and the occupancy $\sigma\left( \mathbf{r} \left( t \right) \right)$ at these points. The predicted pixel colour $\hat{C}\left( \textbf{r} \right)$ in an image is the integration of all the predictions of the sampled points along a ray:
\begin{equation} \label{eq:C_r}
    \hat{C}\left( \textbf{r} \right) =
    \int ^{t_{far}} _{t_{near}} T \left( t \right)
    \sigma \left( \textbf{r} \left( t \right) \right)
    \textbf{c} \left( \textbf{r} \left( t \right), \textbf{d} \right) dt,
\end{equation}
with
\begin{equation}
    T \left( t \right) =
    exp\left( - \int ^{t} _{t_{near}}
    \sigma \left( \textbf{r} \left( s \right) \right) ds \right)
\end{equation}
the accumulated transmittance of light along the ray. This process is repeated for each of the $N_{pixels}$ selected in a rendered image. In this paper, we assume the network is already trained and consider the inverse problem, identifying a camera (or object) pose given a NeRF and a query image.

It is clear that rendering images using the approach above is expensive, requiring $N_{samples}\times N_{pixels}$ network forward passes. NeRFs are typically relatively small networks allowing for ray chunking and batching, but relatively high resolution image rendering can become expensive. This poses challenges for Monte-Carlo or sampling-based localisation schemes.

\subsection{Monte Carlo Localisation}
\label{sec:mcl}

Monte Carlo localisation (MCL) \cite{fox1999monte} is a \textit{particle filter} based localisation method widely used in robot navigation. In a typical MCL scheme, the aim is to localise a robot or camera within an environment with a known map using a finite set of pose samples. To achieve this, the algorithm typically starts by uniformly sampling $N$ particles across the map. The $i$-th particle contains pose information, $\mathbf{X}^i_t$ at a given time instance $t$, such as the robot's position and orientation. The likelihoods $p(\mathbf{Z}_t|\mathbf{X}_t)$ of detecting information $\mathbf{Z}_t$ collected through various sensors on board, given pose particles, are used as particle weights. New particles are then sampled in proportion to these weights, and a motion model $p(\mathbf{X}_t|\mathbf{X}_{t-1})$ can be used to perturb the particles. After a number of iterations, and assuming enough particles are used, the distribution of the particles approximates a probability distribution over the robot state, conditioned on the sensor information history, $p(\mathbf{X}_t|\mathbf{Z}_{1:t})$.

Monte Carlo localisation has the benefit of allowing multi-modal beliefs over poses \cite{fox1999monte,panigrahi2022locstrategies}, and typically converges to a good solution without requiring careful initialisation, but can be computationally expensive, requiring a number of particles to adequately represent beliefs over robot or camera poses.  

Sampling-based localisation schemes like this have been used for NeRF based camera and object \cite{maggio2022locnerf,tian2023multi} localisation, where the map is a trained NeRF implicitly stores the 3D structure and colour of the defined space, and particles are the camera poses $\mathbf{X} = T_{cam} \in SE(3)$, with $SE(3)$ is the Special Euclidean group. In our work, we drop the time-step $t$ and only consider convergence to a static pose. The likelihood $p(\mathbf{Z}|\mathbf{X})$ is constructed by rendering image $\hat{\mathcal{I}}$ at pose $\mathbf{X}$ and comparing this to observed image $\mathbf{Z} = \mathcal{I}_{obs}$.

As mentioned above, rendering an entire image using a NeRF is computationally expensive, so it is cheaper to construct this likelihood by rendering only a subset of pixels or features $\{p_{ij}\}^k$, at row $i$, column $j$, with $k \in (1,N_{pixels})$.

\subsection{Feature Selection}
\label{sec:feature}
Image feature selection has been widely studied in classical computer vision, with conventional wisdom that keypoints or features should ideally be corners, local features for which the signal changes two dimensionally \cite{krystian2002affine}. This intuition underpins the famous Harris corner detector \cite{Harris1988ACC}, which uses an eigenvalue test to look for large variations in intensity in all directions, the SIFT feature detector \cite{lowe2004sift}, which uses a difference of Gaussians approach to find blobs of high contrast, or the FAST 9 feature detector \cite{fast2006edward}, which looks for sharp intensity changes by comparing the intensity of the center pixel  with an outer ring of pixels.  

Oriented FAST is used in the ORB (Oriented FAST and Rotated BRIEF) feature detection and matching scheme \cite{orb2011ethan}, which has proven highly effective in Simultaneous Localisation and Mapping applications \cite{raul2015orbslam}. Despite this preference for corners as feature detectors, prior work has questioned the stability of this choice, with stable region detectors like the Maximally Stable Extremal Region detector (MSER) \cite{mser2004matas} proposed to remedy this. Unlike ORB, MSER tries to find features in more stable image regions. In other words, instead of corners or sharp points, MSER features tend to lie on a continuous patch with similar intensity. Extensive comparisons around repeatability and consistency \cite{mikolyjczyk2005comp} showed MSER produces excellent results, but ORB approaches remain significantly more popular in feature-based visual SLAM applications. This is likely due to the robust matching schemes used within these applications, which arguably negate the effects of the feature detector to a large extent.

\section{Method}
\label{sec:method}

The core hypothesis of this work is that sampling-based localisation using NeRF rendering is highly vulnerable to negative effects introduced by comparing rendered pixels (or patches around these) to these corners or unstable image regions, and that rendering stable features is more robust to these effects. 

We investigate this using the sampling-based localisation scheme described in Algorithm \ref{alg:pose est}. Given an observed image $\mathcal{I}_{obs}$, recall that our goal is to find the camera pose $T_{est} \in SE(3)$ where this image was taken, by selectively rendering image pixels $\widehat{p}_{ij}$ using a pre-trained NeRF network $F_{\theta}$.Ideally we should accomplish this using as few network forward passes as possible, by limiting the number of points along the ray to be rendered, the number of poses queried, and the number of pixels to be rendered. 

\begin{algorithm}
\caption{Camera pose estimation with NeRF}\label{alg:pose est}
\begin{algorithmic}
\State \textbf{Input:} Trained NeRF $F_\theta$, Observed image $\mathcal{I}_{obs}$, Pose range $T_{range}$, Camera intrinsic $K$, Points on the ray $N_{pt}$
\State \textbf{Output:} $N_t$ Best camera pose estimates $T^{est}_{1 \hdots N_t}$
\State Initialise weights $\mathbf{w}$ for all poses
\State Generate candidate pixels $p_{ij}^k, k=\{1 \hdots N_{pixels}\}$ 
\For {$iter = 1 \hdots MaxIter$ }
    \State Sample poses $T_{1 \hdots N_{poses}}$ based on weights $\mathbf{w}$
    \For{$n=1 \hdots N_{poses}$}

        \For{$k=1 \hdots N_{pixels}$}
            \State Get ray $\mathbf{r}_{selected} \gets f(T_n, K, p_{ij}^k, N_{pt})$ 
            \State $\widehat{p}_{ij}^k \gets \hat{C} \left( \mathbf{r}_{selected} \right)$ \Comment{According to \cref{eq:C_r}}
        \EndFor
        \State $e_{ave}^n \gets \mathcal{L}\left( \hat{\mathcal{I}}, \mathcal{I}_{obs} \right)$ \Comment{According to \cref{eq:rgb err}}
    \EndFor

    \State Update $\mathbf{w} \gets \exp \left( -\mathbf{e}_{ave}/\sigma_{e} \right)$ \Comment{According to \cref{eq:weights update}}

\EndFor

\State $T^{est}_{1 \hdots N_t} \gets T^{new}_{1 \hdots N_{poses}}$ with highest associated $\mathbf{w}$

\end{algorithmic}
\end{algorithm}

First, $N_{poses}$ camera poses $T_{1 \hdots N_{poses}}$ are randomly sampled in a defined region $T_{range}$, inferred from the given dataset. For the object-centric localisation problems, the poses are generated using a spherical coordinate pose parametrisation. $N_{poses}$ sets of $\{\theta, \phi, r\}$ are generated within the range of the dataset boundaries. Then they are converted to the $4\times4$ transformation matrices for computation. And for the scene-centric datasets, since it is less amenable to a spherical coordinate system, the pose particles are generated by perturbing the poses in the dataset with various rotation and translations.

In the observed image $\mathcal{I}_{obs}$, a pixel with coordinate $(i, j)$ is denoted by $p_{ij}$. A rendered image is denoted by $\hat{\mathcal{I}}$, and a rendered pixel by $\widehat{p}_{ij}$. The rendering loop is batchified in implementation to maximize GPU utilisation. After rendering, the error between the observed pixel colour and the rendered pixel colour at the same pixel coordinate is calculated. A mean square error (MSE) is used here (mirroring the NeRF training objective), and the error is averaged across RGB channels,
\begin{equation} \label{eq:rgb err}
    e_{ave} = \mathcal{L}\left( \hat{\mathcal{I}}, \mathcal{I}_{obs}  \right) = \dfrac{1}{3}\sum_{c} \dfrac{1}{N_{pixels}} \sum_{i,j} \sqrt{\left\| p_{ij}-\widehat{p}_{ij}\right\| ^{2}},
\end{equation}
where $c \in \{r,g,b\}$ denotes the three colour channels of the pixel, and $i, j$ the selected pixel coordinates.

This error is used to compute a weight for each pose, modelling the likelihood that a set of pixel observations could be made at a given pose,
\begin{equation} \label{eq:weights update}
    \mathbf{w} = \exp \left( -e_{ave}/\sigma_{e} \right),
\end{equation}
where $\sigma_{e}$ is a scalar hyper-parameter, manually tuned to control the weighting ($\sigma_{e}=2$ for all the experiments in this paper). Poses are weighted equally at the first iteration.

\begin{figure*}[!ht]
    \centering
     \includegraphics[width=0.45\textwidth]{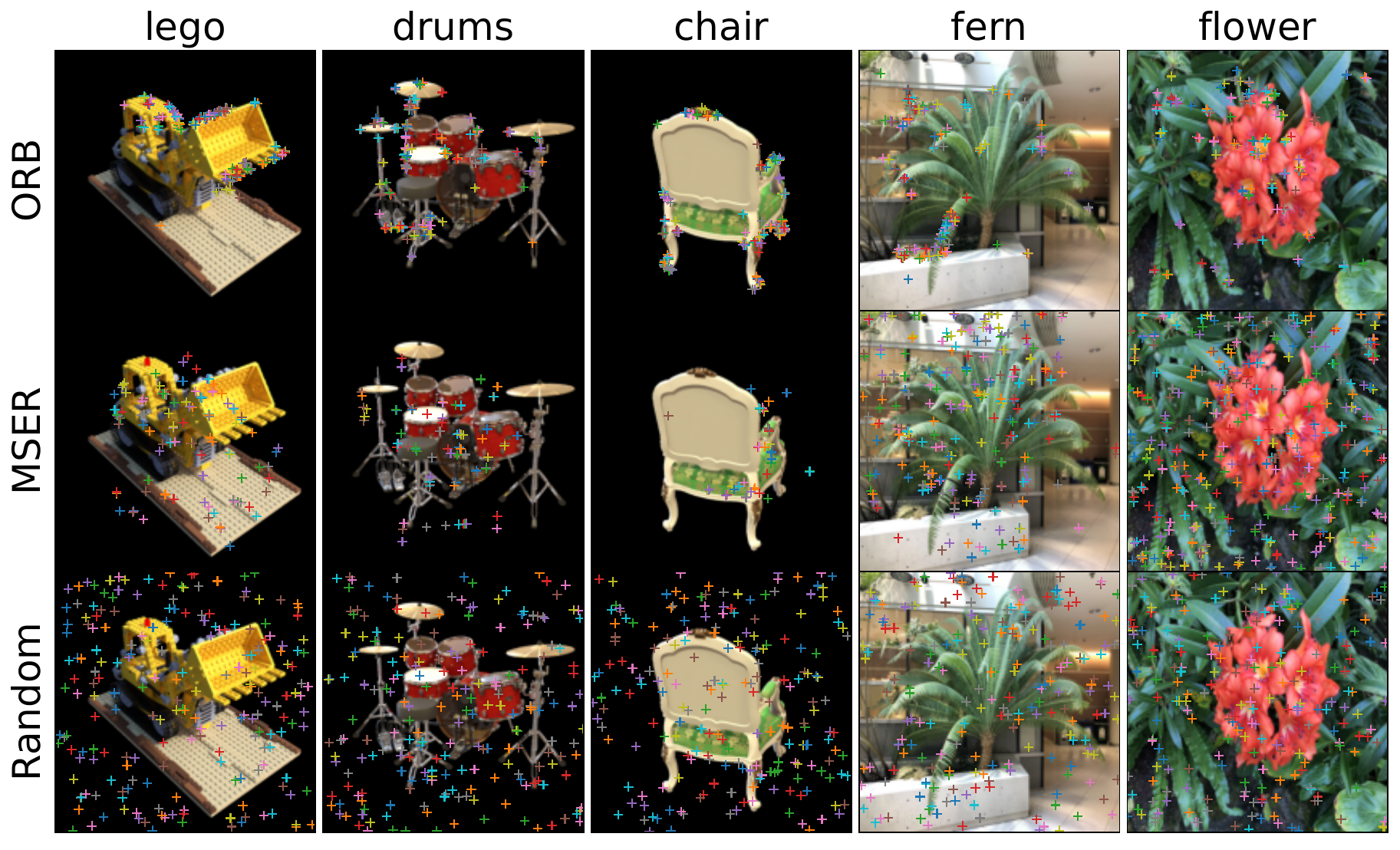}
     \includegraphics[width=0.45\textwidth]{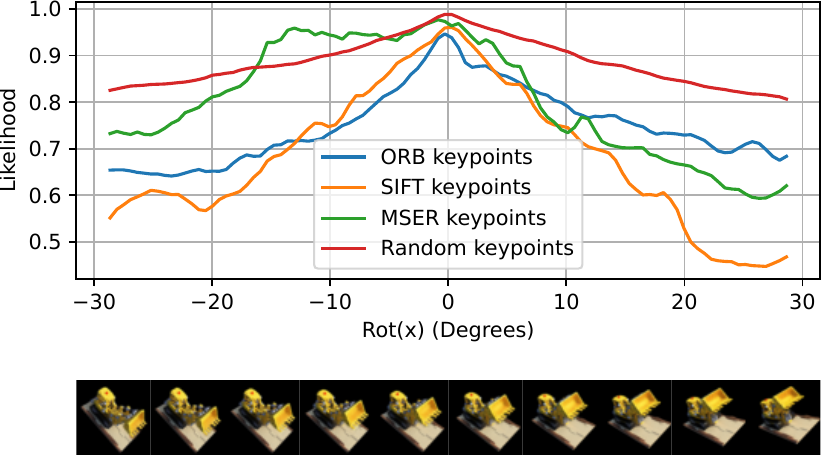}
    \caption{Left: Dataset samples and feature locations. Right: Likelihoods for pose rotation (x) show sharper cut-offs for corner detection schemes. Perturbation along other axes shows similar behaviour. Our hypothesis is that this hinders sampling-based pose estimation scheme convergence.}
    \label{fig:likelihood_hypothesis}
    \vspace{-5mm}
\end{figure*}

We consider two different re-sampling methods. The first relies on the cross entropy method (CEM) \cite{deboer2005cem}. Here, only those particles with the highest weights are kept. These are used to compute a mean pose and associated variance, from which $N_{poses}$ new candidate poses are drawn. Here, the mean, $\mu_{spher}$, and variance, $\sigma^2_{spher}$ of the best poses are calculated, where $spher=\{\theta, \phi, r\}$. New parameters are drawn from a Gaussian with these means and variances.

For scene-centric datasets poses are re-sampled using a pick-and-perturb strategy. Here, the portion of the poses with the highest weights are kept (a third is kept in our experiments). Small random rotation and translation perturbations are then applied to these poses to generate the remaining pose particles, and together the kept and perturbed poses are passed into the next iteration.  In both cases, sampling is repeated iteratively until the maximum number of iterations is reached.

\begin{table*}\centering
\caption{Average percentage pose error}
\label{tab:error}
\scriptsize
\resizebox{0.7\textwidth}{!}{
    \begin{tabular}{lrcccccc}
    \toprule
        & &\multicolumn{5}{c}{\textbf{Dataset}} \\
        \cmidrule{3-7}
        & &lego (\%) &drums (\%) &chair (\%) &fern (\%) &flower (\%) \\
        \cmidrule{1-7}
        \multirow{9}{*}{\textbf{Patch}}
            &ORB &31.42$\pm$3.36 &52.26$\pm$3.80 &40.77$\pm$2.45 &53.68$\pm$4.28 &56.14$\pm$6.20 \\\cmidrule{2-7}
            &ORBrand &22.09$\pm$5.85 &39.81$\pm$5.29 &44.21$\pm$2.12 &57.61$\pm$5.90 &51.74$\pm$8.03 \\\cmidrule{2-7}
            &randfix &23.32$\pm$8.25 &44.07$\pm$7.73 &27.91$\pm$5.64 &31.87$\pm$5.25 &\textbf{24.31$\pm$5.30} \\\cmidrule{2-7}
            &rand &13.39$\pm$6.28 &26.17$\pm$7.53 &28.63$\pm$5.87 &26.76$\pm$4.37 &31.87$\pm$8.92 \\\cmidrule{2-7}
            &MSER &13.23$\pm$5.93 &21.36$\pm$6.27 &18.62$\pm$6.78 &28.01$\pm$4.12 &42.11$\pm$6.07 \\\cmidrule{2-7}
            &MSERrand &\textbf{12.23$\pm$5.74} &\textbf{14.69$\pm$5.57} &\textbf{15.45$\pm$5.73} &\textbf{22.25$\pm$5.85} &37.64$\pm$7.48 \\
        \cmidrule{1-7}
        \multirow{9}{*}{\textbf{Pixel}}
            &ORB &37.26$\pm$2.89 &51.40$\pm$3.50 &42.23$\pm$2.05 &61.26$\pm$4.05 &49.15$\pm$5.10 \\\cmidrule{2-7}
            &ORBrand &25.01$\pm$5.51 &44.61$\pm$3.04 &43.44$\pm$2.03 &62.02$\pm$6.30 &48.69$\pm$6.43 \\\cmidrule{2-7}
            &randfix &12.49$\pm$5.32 &15.61$\pm$5.35 &22.55$\pm$5.42 &31.28$\pm$4.52 &\textbf{22.26$\pm$4.78} \\\cmidrule{2-7}
            &rand &12.17$\pm$6.10 &17.46$\pm$7.05 &27.95$\pm$6.42 &23.08$\pm$3.04 &26.27$\pm$7.25 \\\cmidrule{2-7}
            &MSER &12.17$\pm$5.50 &\textbf{13.83$\pm$5.34} &14.71$\pm$5.89 &24.70$\pm$3.52 &39.18$\pm$8.67 \\\cmidrule{2-7}
            &MSERrand &\textbf{10.31$\pm$5.22} &15.41$\pm$6.02 &\textbf{11.90$\pm$5.34} &\textbf{19.20$\pm$4.07} &38.22$\pm$8.48 \\
    \bottomrule
    \end{tabular}}
\end{table*}

\section{Experiments}
\label{sec:exp}
Our core goal is to reduce the number of forward passes required to infer pose using the sampling-based localisation schemes described above while maintaining good pose estimation quality. We evaluate a range of different parameters and settings on a number of different datasets to investigate ways of accomplishing this.

The experiments conducted here consider camera pose estimation in static scenes only, although are generalisable to moving cameras with the addition of motion models. Both object-centric (lego, drums, chair) \cite{mildenhall2020nerf} and scene-centric (flower, fern) \cite{mildenhall2019llff} datasets are considered. Object-centric datasets use synthetic images generated with the origin of the coordinate set at the bottom center of an object, with camera poses known to be around this fixed object, facing towards the origin. Real-world scenes are captured using outward facing cameras, within some limited region specified by a set of training poses.

Our experiments assume that the NeRF network $F_\theta$ is pre-trained for each dataset and the weights are not updated during any of the experiments. For consistency across the experiments, the pose estimation process is run for 20 iterations for all parameter choices and experiments. The number is chosen to ensure most runs would converge to a reasonable estimate. The key parameters investigated in this work include the number of particles, or poses to be sampled, $N_{poses}$; the number of pixels to be selected, $N_{pixels}$; the number of points, $N_{pts}$, along the ray to be used for rendering (render quality); the pixel selection method (random, ORB or MSER); likelihoods based on individual pixels or a $3\times3$ patch around each selected pixel; and randomized pixel selection over iterations vs fixed pixels.

Pose estimation runs for 20 iterations across multiple parameter combinations and pose error is evaluated by comparing points transformed by ground truth and estimated poses. Experiments were conducted using 32x2.4GHz CPUs, with networks run on a 2080 Ti GPU. This allowed chunking and batch sizes of approximately 30k forward pass. These experiments are time consuming, with larger pose and pixel numbers requiring up to 40 minutes to return a single pose estimate.

It is clear that reducing the number of pixels and sampled poses reduces the number of forward passes, although too few may affect localisation convergence. However, the effects of the remaining parameters are less clear. In particular, when it comes to the pixel selection method, our hypothesis is that rendering corners may be less effective when it comes to localisation, as these produce unstable likelihoods that cut off sharply or are more vulnerable to rendering errors on the edges or corners of objects.
To illustrate this, Figure \ref{fig:likelihood_hypothesis} shows keypoint samples using different methods, and the likelihood calculated in \cref{eq:weights update} for 1000 pixels sampled from a $100\times 100$ pixel test image (tiny NeRF data), for images generated as the query pose is rotated away from the test image pose ($0^\circ$) along the x-axis. Keypoint selection strategies are more dependent on the image content than random pixel selection, with corner detectors (ORB, SIFT) cutting off more rapidly than MSER detection. Random pixel selection is stable with higher numbers of pixels, but degrades with fewer. We speculate that the wider cut-off in likelihood attributed to more stable detection strategies is better suited to sampling-based localisation with fewer poses, and is likely to aid convergence.

\section{Results and Discussion}
\label{sec:results}

\subsection{Metrics}

To provide a common measure across the object facing and forward scene facing datasets, which have varying scales, we report percentage error in our results. We select the best 5 estimated poses in the last sampling iteration and evaluate the average error. This is consistent with the CEM pose estimate (a set of particles are retained).

The error is first evaluated by comparing points transformed by ground truth and estimated poses. Here, a test point $p_{test} \in R^3$ is chosen ($[1,1,1]$ in our implementation), and the ground truth camera pose $T_{gt}$ and the 5 highest weighted camera particles $T^{est}_{1 \hdots 5}$ are applied to this test point, hence both translation and rotation error are taken into account in a single metric. We report the average Euclidean distance between the transformed ground truth test point and corresponding 5 transformed estimation points. The percentage error is scaled by the maximum distance in the dataset.\footnote{12 for the synthetic object facing dataset and 2 for the forward facing dataset.}

In order to get more insight into the effects of the feature selection method, translational and rotational error (TE and RE) are also provided for a selection of experiments. TE reflects the average displacement error between the ground truth and the estimated camera position, scaled by the maximum distance in the dataset. RE reflects the viewing angle difference between the target and estimated pose, scaled to a percentage of 180\degree (hence 2.78\% is equivalent to converging to less than 5\degree error). The results in Table \ref{tab:TE_RE} are provided under this metric.

\subsection{Results}

Table \ref{tab:error} shows the average localisation error across experiments\footnote{Averaged over multiple runs with $N_{pixels} = (100, 500, 1000)$, $N_{pts} = (16, 32, 64)$, and $N_{poses} = (5, 10, 15, 20, 25, 30, 35, 40, 45, 50, 100)$} with associated 95\% confidence intervals. Since these results are aggregated, they depend solely on the point selection method and pixel rendering strategy.

Here, `ORBrand' and `MSERrand' generate a large number of initial features and pixels are randomly selected from this pool of candidate features in each iteration. `Rand' randomly select pixels across the entire image in each iteration. `ORB', `randfix' and `MSER' select $N_{pixels}$ initial pixel locations, which do not change across iterations. Patch and pixel experiments used an equivalent number of rendered pixels (a fixed number of network forward passes), and it is clear that rendering patch regions reduces the information conveyed by the rendering likelihood, inhibiting convergence.

Across most datasets, the MSERrand feature selection method produces the best result, significantly outperforming the ORB selection method. This strongly supports our hypothesis that rendering stable features improves NeRF localisation.

\begin{figure}
    \centering
    \includegraphics[width=\linewidth]{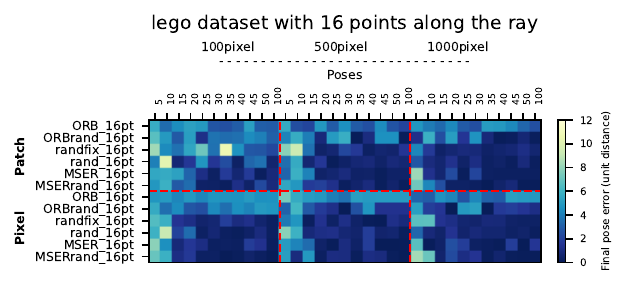}
    \includegraphics[width=\linewidth]{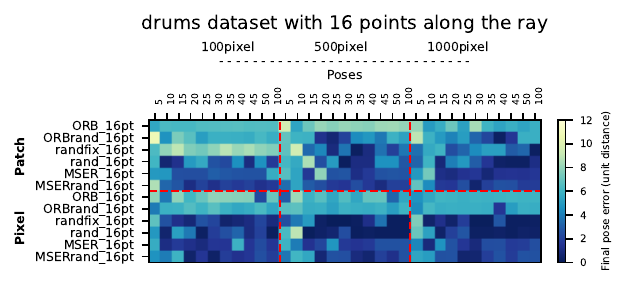}
    \includegraphics[width=\linewidth]{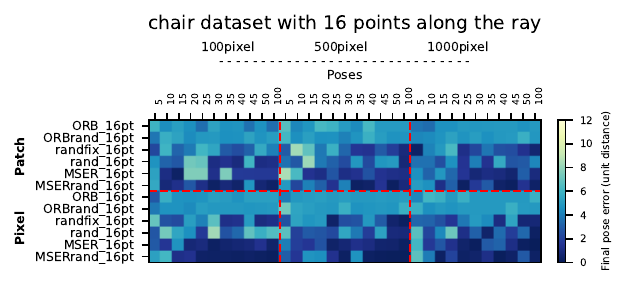}
    \includegraphics[width=\linewidth]{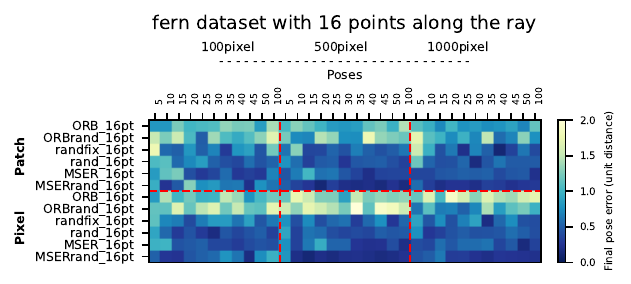}
    \includegraphics[width=\linewidth]{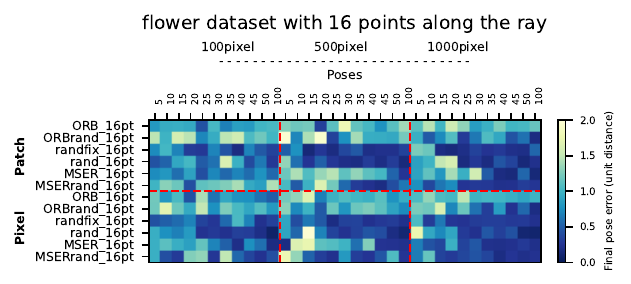}
    \caption{Final pose estimation errors when varying rendered pixels and pose samples (16 points rendered along NeRF rays).}
    \label{img:16pt}
\end{figure}

\begin{figure}
\centering
    \includegraphics[width=\linewidth]{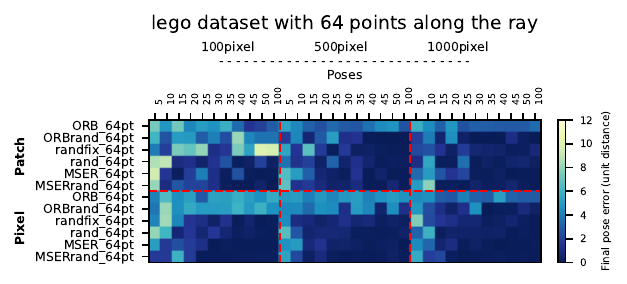}
    \includegraphics[width=\linewidth]{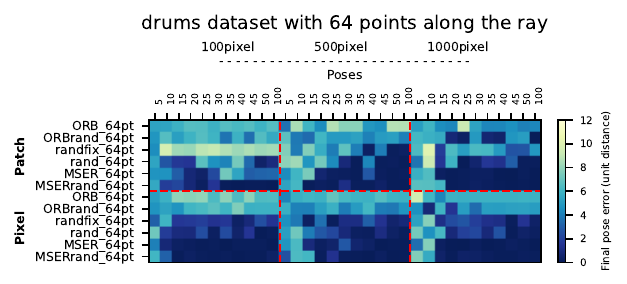}
    \includegraphics[width=\linewidth]{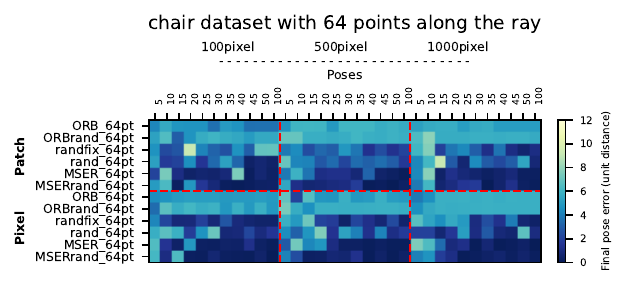}
    \includegraphics[width=\linewidth]{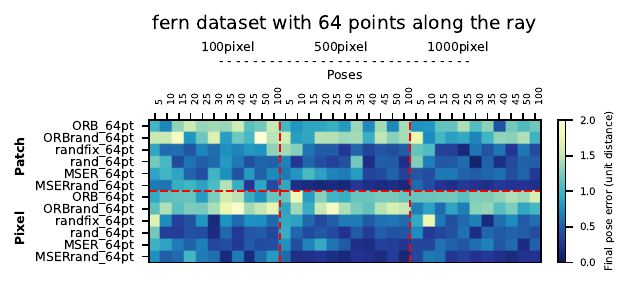}
    \includegraphics[width=\linewidth]{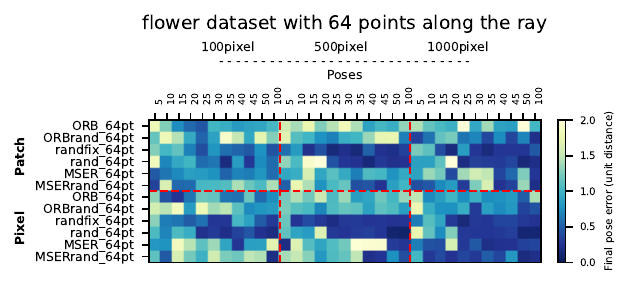}
    \caption{Final pose estimation errors when varying rendered pixels and pose samples (64 points rendered along NeRF rays). }
    \label{img:64pt}
\end{figure}

\begin{table*}\centering
\caption{TE and RE of the experiment with 100 pixels and 45 poses}
\label{tab:TE_RE}
\scriptsize
\resizebox{0.75\textwidth}{!}{
    \begin{tabular}{lrcccccccccc}
    \toprule
        & &\multicolumn{5}{c}{\textbf{Translation Error (\%)}} &\multicolumn{5}{c}{\textbf{Rotation Error (\%)}} \\\cmidrule{3-12} 
        & &lego &drums &chair &fern &flower &lego &drums &chair &fern &flower \\
        \cmidrule{1-12}
        \multirow{9}{*}{\textbf{Patch}}
            &ORB &18.99 &35.19 &32.04 &27.34 &23.69 &21.03 &49.72 &42.57 &21.19 &13.91 \\\cmidrule{2-12}
            &ORBrand &24.09 &27.20 &32.19 &20.99 &18.24 &28.12 &31.89 &43.58 &14.60 &21.15 \\\cmidrule{2-12}
            &randfix &45.04 &50.32 &18.91 &21.29 &14.74 &43.62 &60.68 &18.42 &15.87 &8.54 \\\cmidrule{2-12}
            &rand &6.22 &19.45 &14.42 &23.12 &22.83 &4.95 &15.83 &12.68 &4.19 &\textbf{5.18} \\\cmidrule{2-12}
            &MSER &4.37 &17.41 &\textbf{4.18} &21.47 &\textbf{10.84} &2.94 &16.29 &\textbf{3.56} &6.76 &10.23 \\\cmidrule{2-12}
            &MSERrand &\textbf{1.28} &\textbf{9.25} &13.16 &\textbf{14.29} &22.38 &\textbf{1.10} &\textbf{7.31} &17.02 &\textbf{2.57} &19.98 \\
        \cmidrule{1-12}
        \multirow{9}{*}{\textbf{Pixel}}
            &ORB &22.94 &38.43 &29.64 &22.37 &21.20 &25.48 &57.36 &37.80 &13.10 &13.51 \\\cmidrule{2-12}
            &ORBrand &33.42 &30.70 &32.12 &25.66 &16.02 &41.89 &37.02 &42.34 &16.53 &17.92 \\\cmidrule{2-12}
            &randfix &4.60 &5.68 &15.77 &42.01 &\textbf{12.80} &3.07 &4.72 &14.93 &4.97 &10.56 \\\cmidrule{2-12}
            &rand &1.34 &5.30 &23.93 &26.17 &14.86 &1.16 &4.24 &21.86 &3.32 &\textbf{4.94} \\\cmidrule{2-12}
            &MSER &4.17 &4.64 &3.76 &23.75 &23.57 &3.52 &2.92 &3.40 &2.80 &17.63 \\\cmidrule{2-12}
            &MSERrand &\textbf{1.23} &\textbf{3.86} &\textbf{1.12} &\textbf{8.35} &20.52 &\textbf{1.05} &\textbf{2.45} &\textbf{1.01} &\textbf{1.44} &17.47 \\
    \bottomrule
    \end{tabular}
    }
\end{table*}

Figure \ref{img:16pt} and Figure \ref{img:64pt} show the final pose estimation error for each dataset after 20 iterations when rendering 16 and 64 points along the ray respectively, with little difference between these.\footnote{Experiments with 32 points show similar patterns.} In each heatmap, the x-axis shows varying numbers of initial pose particles with increasing pixel numbers rendered at each pose. The y-axis shows the different pixel sampling methods, with the same order as presented in the tables. Red lines separate experimental groupings for readability.

Similar performance holds when evaluating the translation and rotation errors separately. Table \ref{tab:TE_RE} presents the evaluation of one of the experiment settings when rendering 100 pixels and using 45 initial pose particles. This setting appears to provide the best balance between convergence and requiring a relatively low number of forward passes to complete. Results shows that pose estimation with MSER sampling method generally performs better across most datasets. When comparing with the performance of ORB sampling, the error percentage drops significantly. Similar trends are seen across other pixel and pose numbers.  

As expected, rendering more pixels and sampling more initial poses generally results in better estimation. In general, results improve in the order of ORB, ORBrand, randfix, rand, MSER and MSERrand. This corresponds to moving from less stable corners to random points (stable in expectation) to points in stable regions. Adding random variations to pixel selection across iterations further improves convergence. In easier datasets like lego and drums, random and MSER detection-based runs provide good estimation with as few as 15 initial poses.

Both the tables and the figures show that for ORB detection, rendering patches performs slightly better than rendering single pixels, aligning with iNeRF \cite{yen2020inerf} findings that rendering interest regions increases performance. This also supports our hypothesis since rendering (observing) a larger area around the corner points smooths the error likelihood.

MSER did not perform as well on the flower dataset, where random selection was more effective. This may be because the images in the flower dataset have many self-similar areas of red petals and green leaves (as shown in Figure \ref{fig:likelihood_hypothesis}), which are less informative in stable regions. Random pixel selection is less vulnerable to this.

In addition, Table \ref{tab:forward_pass} shows the minimum number of network forward passes required to reach $\le 10\%$ pose error for each dataset and method. The fern and flower datasets are quite difficult to localise in, as convergence in these complex real scenes with scattered features is hindered by the natural and amorphous objects therein. Table entries without a number mean none of the runs of this feature selection method converged to under $\le 10\%$ error. This is a relatively noisy set of measurements, as it does not consider aggregate performance over multiple runs (Table \ref{tab:error} is a more robust measure), but does give some indication of the orders of magnitude required for convergence. In general, with a sensible feature selection strategy, it is possible to reduce the number of network forward passes roughly tenfold.

\begin{table}\centering
    \caption{Forward passes to achieve $\le 10\%$ pose error}
    \label{tab:forward_pass}
    \scriptsize
    \resizebox{0.48\textwidth}{!}{
        \begin{tabular}{lrccccc}
        \toprule
            & &lego &drums &chair &fern &flower \\
            \cmidrule{1-7}
            \multirow{9}{*}{\textbf{Patch}}
                &ORB      &- &- &- &- &- \\\cmidrule{2-7}
                &ORBrand  &6.08m &8.4m &5.28m &- &31.70m \\\cmidrule{2-7}
                &randfix  &3.73m &5.92m &720k &29.3m &\textbf{1.25m} \\\cmidrule{2-7}
                &rand     &\textbf{744k} &\textbf{720k} &1.01m &20.2m &4.24m\\\cmidrule{2-7}
                &MSER     &1.17m &3.75m &\textbf{480k} &- &12.5m \\\cmidrule{2-7}
                &MSERrand &1.92m &1.28m &512k &\textbf{7.89m} &7.20m \\
            \cmidrule{1-7}
            \multirow{9}{*}{\textbf{Pixel}}
                &ORB      &- &- &- &- &- \\\cmidrule{2-7}
                &ORBrand  &13.76m &23.04m &- &- &21.70m \\\cmidrule{2-7}
                &randfix  &\textbf{512k} &1.34m &\textbf{816k} &7.30m &3.85m \\\cmidrule{2-7}
                &rand     &624k &\textbf{528k} &896k &\textbf{579k} &1.54m \\\cmidrule{2-7}
                &MSER     &1.54m &672k &1.08m &24.3m &\textbf{496k}  \\\cmidrule{2-7}
                &MSERrand &1.43m &1.63m &880k &3.60m &12.5m \\
        \bottomrule
        \end{tabular}
    }\vspace{-5mm}
\end{table}

Note that none of the ORB selection method ever converged to below 10\% estimation error. Most of the runs with few forward passes are from runs that render 16 or 32 points along the ray, with around 30 initial poses on average. This indicates that when choosing between rendering high quality image pixels, for greater coverage of space, or smartly selecting the pixel to render, the latter plays a more important role for localisation performance.

\section{Limitations and Future Work}
\label{sec:limitations}

As mentioned above, the experiments conducted here are time consuming due to the large number of forward passes required for localisation. This limited our investigations to the pixel loss common in NeRF literature when constructing likelihoods. It may be that alternative strategies leveraging classical feature matching approaches or descriptors as patch likelihoods (eg. SSIM \cite{ssim}) would be more effective, although these would require additional compute. Similarly, losses that make better use of patches (eg. SSIM) may be worth exploring, although rendering larger patches is likely to be too expensive for practical use. Our current experiments have also assumed static scenes and cameras. However, we expect that generalising to moving cameras would not change these results significantly. 

\section{Conclusion}
\label{sec:conclusion}
This paper investigated strategies to speed up sampling-based localisation with NeRFs. Our core finding is that rendering sampled stable features, where image spatial changes are less frequent, outperforms rendering approaches relying on corners or interest points and regions surrounding these, where the feature changes rapidly. Results also show that reducing the number of points to be integrated along a NeRF ray does not significantly inhibit localisation, with rendered pixels and the number of poses sampled playing a more important role.

%% This section was initially prepared using BibTeX.  The .bbl file was
%% placed here later
%\bibliography{publications}
%\bibliographystyle{named}
%% The file named.bst is a bibliography style file for BibTeX 0.99c

\bibliographystyle{IEEEtran}
\bibliography{references}
\end{document}